\documentclass[letterpaper, 10 pt, conference]{ieeeconf}  

\IEEEoverridecommandlockouts                              

\usepackage{amsmath}
\usepackage{amssymb}
\usepackage{booktabs}
\usepackage{multirow}
\usepackage{graphicx}
\usepackage[table]{xcolor}
\usepackage{svg}
\usepackage{cite}

\usepackage[hidelinks]{hyperref}

\definecolor{bestcolor}{RGB}{139,190,235}
\definecolor{secondcolor}{RGB}{218,233,247}
\definecolor{egoColor}{RGB}{233,113,50}
\definecolor{actorColor}{RGB}{21,96,130}
\title{\LARGE \bf
RECAST: From Log Replay to Closed-Loop Driving Simulation
with View-Complete Actors
}

\author{
Zijun Zhao$^{1}$,
Liewen Liao$^{1}$,
Kang Shen$^{1}$,
Songan Zhang$^{1,\dagger}$, and
Ming Yang$^{2}$%
\thanks{$^{1}$Zijun Zhao, Liewen Liao, Kang Shen, and Songan Zhang are with the
Global Institute of Future Technology, Shanghai Jiao Tong University,
Shanghai 200240, China.}%
\thanks{$^{2}$Ming Yang is with the School of Automation and Intelligent Sensing,
Shanghai Jiao Tong University, and the Key Laboratory of System Control and
Information Processing, Ministry of Education of China,
Shanghai 200240, China.}%
\thanks{$^{\dagger}$Corresponding author: Songan Zhang
(\texttt{songanz@sjtu.edu.cn}).}%
}

\begin{document}

\maketitle
\thispagestyle{empty}
\pagestyle{empty}

\begin{abstract}
Closed-loop driving simulation requires rendered observations to remain reliable
as the ego vehicle and surrounding actors move beyond their recorded
trajectories, exposing views absent from the source log. Existing data-driven
simulators reconstruct dynamic actors from sparse observations, which can
result in rendering artifacts under these viewpoint changes. We introduce RECAST
(REconstructing Controllable Actors for Simulation and Testing), a 3D Gaussian
Splatting framework that generates a view-complete actor from a single
segmented vehicle observation in a driving log and registers the generated
actor in the reconstructed scene. RECAST supports planner-in-the-loop rendering under
controlled ego--actor interactions. To adapt an image-to-3D prior to real vehicles,
we further introduce RECAR, a dataset of approximately 20K real vehicles with
600K background-free RGBA images spanning diverse vehicle colors and types.
We use two-stage adaptation to improve vehicle generation from real driving-log observations.
At the actor level, RECAST reduces
$\mathrm{FD}_{\mathrm{incep}}$ from 9.788 to 7.992 relative to unadapted
TRELLIS. At the scene level, under actor motion beyond logged trajectories,
RECAST reduces $\mathrm{FD}_{\mathrm{incep}}$ from 129.35 to 112.10 and increases
$\mathrm{CLIP}_{\mathrm{margin}}$ ($\times1000$) from 0.14 to 3.47 relative to
Street Gaussians. We demonstrate planner-in-the-loop simulation with the
image-conditioned planner GTRS-Dense. Compared with
native Street Gaussians actors, RECAST increases the no-collision (NC) rate from 22.2\% 
(12/54) to 63.0\% (34/54) and the mean minimum predicted time-to-collision (TTC) from 0.798 s 
to 2.150 s. These experiments show that RECAST supports closed-loop
planner evaluation under controlled ego--actor interactions beyond log replay.
Project page: https://zijunkr.github.io/RECAST/
\end{abstract}

\section{Introduction}

Neural radiance fields and 3D Gaussian Splatting
enable photorealistic novel-view synthesis.
Their extensions to driving scenes support data-driven simulation for
autonomous driving~\cite{ad3d_survey,nerf,3dgs,
mars,unisim,neurad,omnire,hugsim,streetgaussians}.
These reconstruction methods better preserve the appearance and geometry of
real environments than engine-based simulators such as CARLA~\cite{carla}.
However, reconstructed driving scenes are typically evaluated under log replay or small 
camera translations around recorded viewpoints,
where the ego vehicle and surrounding actors remain close to their recorded
states.

\begin{figure}[!t]
\centering
\includegraphics[width=\columnwidth]{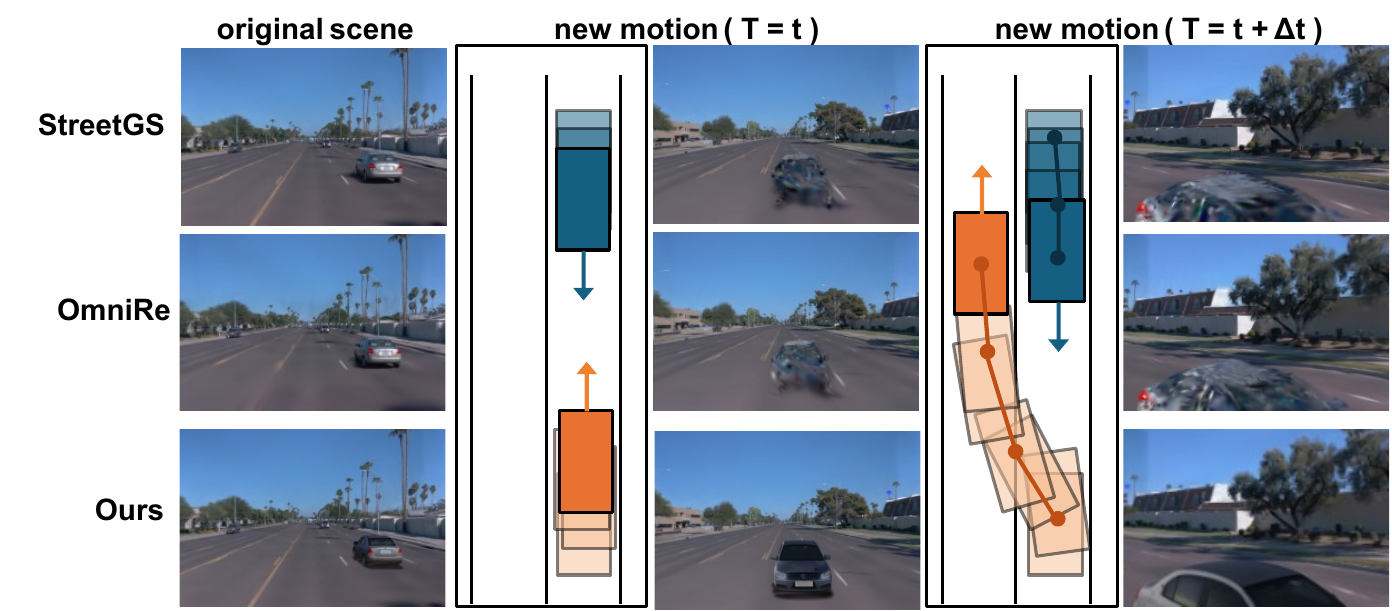}
\caption{\textbf{Actor appearance beyond log replay.} Actor motion beyond the 
logged trajectory exposes previously unseen sides of the 
vehicle, revealing rendering artifacts in Street Gaussians (StreetGS)~\cite{streetgaussians} and OmniRe~\cite{omnire}. RECAST 
generates a view-complete actor from a logged
image and preserves coherent vehicle appearance under these viewpoint changes.
\quad\mbox{\protect\textcolor{egoColor}{\protect\rule{0.8em}{0.8em}}~Ego\quad
\protect\textcolor{actorColor}{\protect\rule{0.8em}{0.8em}}~Actor}
}
\label{fig:motivation}
\vspace{-4mm}
\end{figure}

Street Gaussians (StreetGS)~\cite{streetgaussians} and OmniRe~\cite{omnire} reconstruct
dynamic driving scenes by modeling tracked vehicles separately from the
static background. This separation supports actor motion within a reconstructed
scene while retaining vehicle appearance from driving logs. However, sparse
logged observations provide limited coverage of each vehicle. When ego and
actor motion depart from the recorded trajectories, the changing relative
viewpoints expose unseen sides of the vehicle. Figure~\ref{fig:motivation}
illustrates the resulting artifacts in Street Gaussians and OmniRe under
these viewpoint changes, motivating view-complete actors for interactive
and planner-in-the-loop simulation.

AutoSplat~\cite{autosplat} uses vehicle templates and symmetry priors to improve actor
reconstruction, but still relies on observations of the
target vehicle. HUGSIM~\cite{hugsim} and MADrive~\cite{madrive} obtain broader view coverage by inserting
view-complete assets from external libraries, replacing
the target vehicle with an independently captured asset.

Recent image-to-3D methods further make it possible to recover vehicle assets
from sparse real-world observations. Drive-1-to-3~\cite{drive1to3}, Asset Harvester~\cite{asset_harvester}, 3DCarGen~\cite{3dcargen} and
Unposed-to-3D~\cite{unposed_to_3d} demonstrate promising generation and novel-view synthesis
quality for real vehicles. However, these methods can
produce unrealistic vehicles at viewpoints absent from the source log.
These artifacts can interfere with the planner's visual input during
closed-loop simulation.

To address this problem, we present \textbf{RECAST}, a closed-loop driving
simulation framework with view-complete actors generated from logged images.
Given a single segmented observation of a surrounding vehicle, RECAST
generates a view-complete actor that retains appearance cues from the observed vehicle.
RECAST then registers the generated actor in the reconstructed scene.
The resulting scene supports controllable ego and actor motion and renders
observations from the current simulation state for planner-in-the-loop
execution. To improve vehicle generation from sparse observations in real
driving logs, we further
introduce \textbf{RECAR}, a large-scale real-vehicle dataset used to adapt the
underlying image-to-3D prior to the real driving domain.

Our contributions are as follows:
\begin{itemize}
\item \textbf{Closed-loop simulation with the controllable actor:}
RECAST integrates the generated view-complete vehicle actor into reconstructed real-world
scenes, enabling controlled ego--actor interactions and planner-in-the-loop execution.
\item \textbf{Vehicle generation from sparse observations:}
We introduce \textbf{RECAR}, with approximately 20K real vehicles and 600K
background-free RGBA images, and a two-stage adaptation strategy for generating
a view-complete actor from a single segmented driving-log image.
\item \textbf{Closed-loop planner evaluation:}
We demonstrate planner-in-the-loop evaluation with an
image-conditioned planner and quantify how actor representation
affects closed-loop driving outcomes.
\end{itemize}

\section{RELATED WORK}
\subsection{3D Vehicle Generation}
Methods for 3D generation can be broadly grouped
into novel-view generation and direct 3D generation. Novel-view synthesis
methods, including Zero-1-to-3~\cite{zero123}, SyncDreamer~\cite{syncdreamer},
Wonder3D~\cite{wonder3d}, and SV3D~\cite{sv3d}, synthesize unobserved views from
sparse inputs but require an additional lifting stage to obtain manipulable 3D
actors. General-purpose image-to-3D methods, such as Real3D~\cite{real3d},
LN3Diff~\cite{ln3diff}, 3DTopia-XL~\cite{3dtopia-xl}, SAR3D~\cite{sar3d},
TRELLIS~\cite{trellis}, and Hunyuan3D 2.0~\cite{hunyuan3d}, directly predict
structured 3D representations. These methods are typically trained on
multiple object categories and are not explicitly adapted to sparse
observations from real driving logs.

Vehicle-specific approaches introduce stronger domain priors. Drive-1-to-3
~\cite{drive1to3} and DreamCar~\cite{dreamcar} incorporate vehicle priors for
novel-view synthesis or prior-guided reconstruction, while Asset
Harvester~\cite{asset_harvester} and 3DCarGen~\cite{3dcargen} synthesize
multi-view observations before lifting them into 3D. Unposed-to-3D
~\cite{unposed_to_3d} directly predicts metric-scale vehicle Gaussians from
real-world images. RECAST instead adapts a structured image-to-3D prior with
vehicle-specific supervision to generate view-complete actors from single
observations in real driving logs and integrate the generated actors into
controlled simulation.

\subsection{Driving Scene Reconstruction and Simulation}
Driving scene reconstruction recovers scene geometry and appearance from
driving logs for novel-view rendering. NeRF-based
methods, including Neural Scene Graphs~\cite{nsg}, MARS~\cite{mars},
UniSim~\cite{unisim}, and NeuRAD~\cite{neurad}, model dynamic urban scenes using
neural radiance fields, while recent 3DGS approaches, such as
DrivingGaussian~\cite{drivinggaussian}, Street Gaussians
~\cite{streetgaussians}, and OmniRe~\cite{omnire}, represent the static background and 
individual actors with separate sets of 3D Gaussians, enabling efficient rendering.

These methods preserve the identity of surrounding actors, but their actor
representations are optimized from the observations available in the original
driving sequence, leaving unseen vehicle surfaces weakly constrained.
AutoSplat~\cite{autosplat} mitigates this issue through template initialization and
reflected Gaussian consistency, although template and symmetry priors provide
limited cues for instance-specific appearance on unobserved surfaces.

View-complete asset insertion takes the complementary approach. HUGSIM
~\cite{hugsim} inserts non-native vehicles reconstructed from dense
\(360^\circ\) observations in 3DRealCar~\cite{3drealcar}, while MADrive
retrieves visually similar vehicles from MAD-CARS~\cite{madrive}. Such
approaches provide complete actor geometry and appearance under large relative
viewpoint changes, but replace the target actor with an independently
captured vehicle. RECAST instead generates a view-complete actor from an observation
of the target vehicle for controlled closed-loop simulation.

\section{METHOD}
\subsection{Framework Overview}
\label{sec:framework_overview}

\begin{figure*}[!t]
    \centering
    \includegraphics[width=1\linewidth]{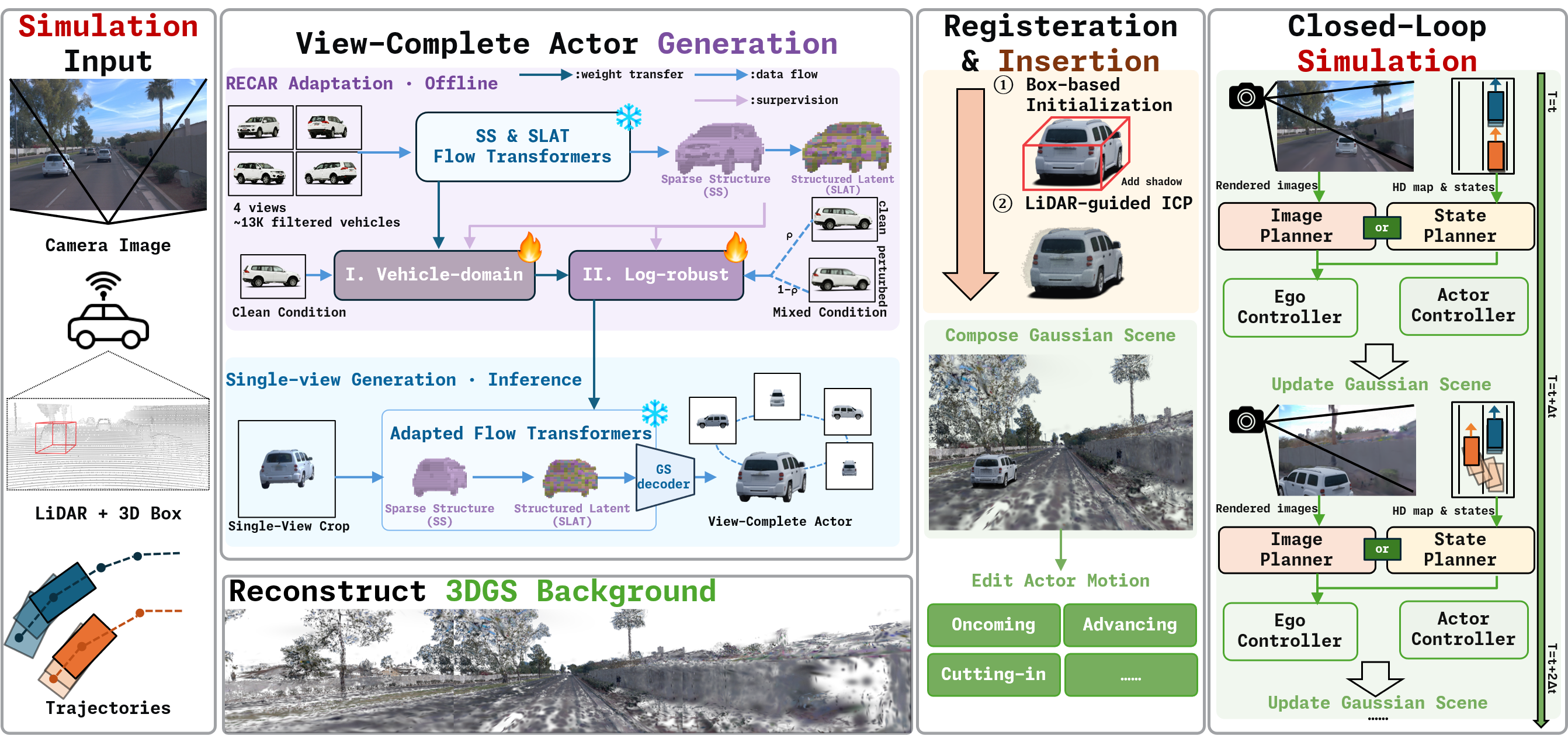}
    \caption{\textbf{RECAST overview.} Multi-view pseudo-targets from RECAR support
    vehicle-domain and driving-log-robust (log-robust) adaptation. Given a single segmented
    driving-log image, the adapted model generates a view-complete Gaussian actor,
    which is registered using the logged 3D box and LiDAR points and then inserted into
    the reconstructed background. During closed-loop simulation, the ego controller
    executes the planner's predicted trajectory, while the actor controller follows
    the specified motion. Updated states and camera poses provide the next planner input.
    \quad\mbox{\protect\textcolor{egoColor}{\protect\rule{0.8em}{0.8em}}~Ego\quad
\protect\textcolor{actorColor}{\protect\rule{0.8em}{0.8em}}~Actor}
    }
    \vspace{-4mm}
    \label{fig:recast_overview}
\end{figure*}

Figure~\ref{fig:recast_overview} summarizes RECAST. Given a driving log,
Street Gaussians~\cite{streetgaussians} reconstructs the scene and separates
a surrounding vehicle selected as the target actor. RECAST generates a
view-complete representation of the target actor from one segmented observation,
registers the generated Gaussian actor using the logged 3D box and LiDAR points,
and combines the registered actor with the reconstructed background. During closed-loop 
simulation, the ego controller executes the planner's predicted trajectory, while the 
actor controller updates the target actor's state.

\subsection{View-Complete Actor Generation}
\label{sec:actor_generation}
Given a single actor observation, we generate a view-complete 3D Gaussian
actor. We adapt TRELLIS~\cite{trellis} using RECAR through vehicle-domain
and driving-log-robust adaptation.

\begin{figure}[!t]
    \centering
    \includegraphics[width=\linewidth]{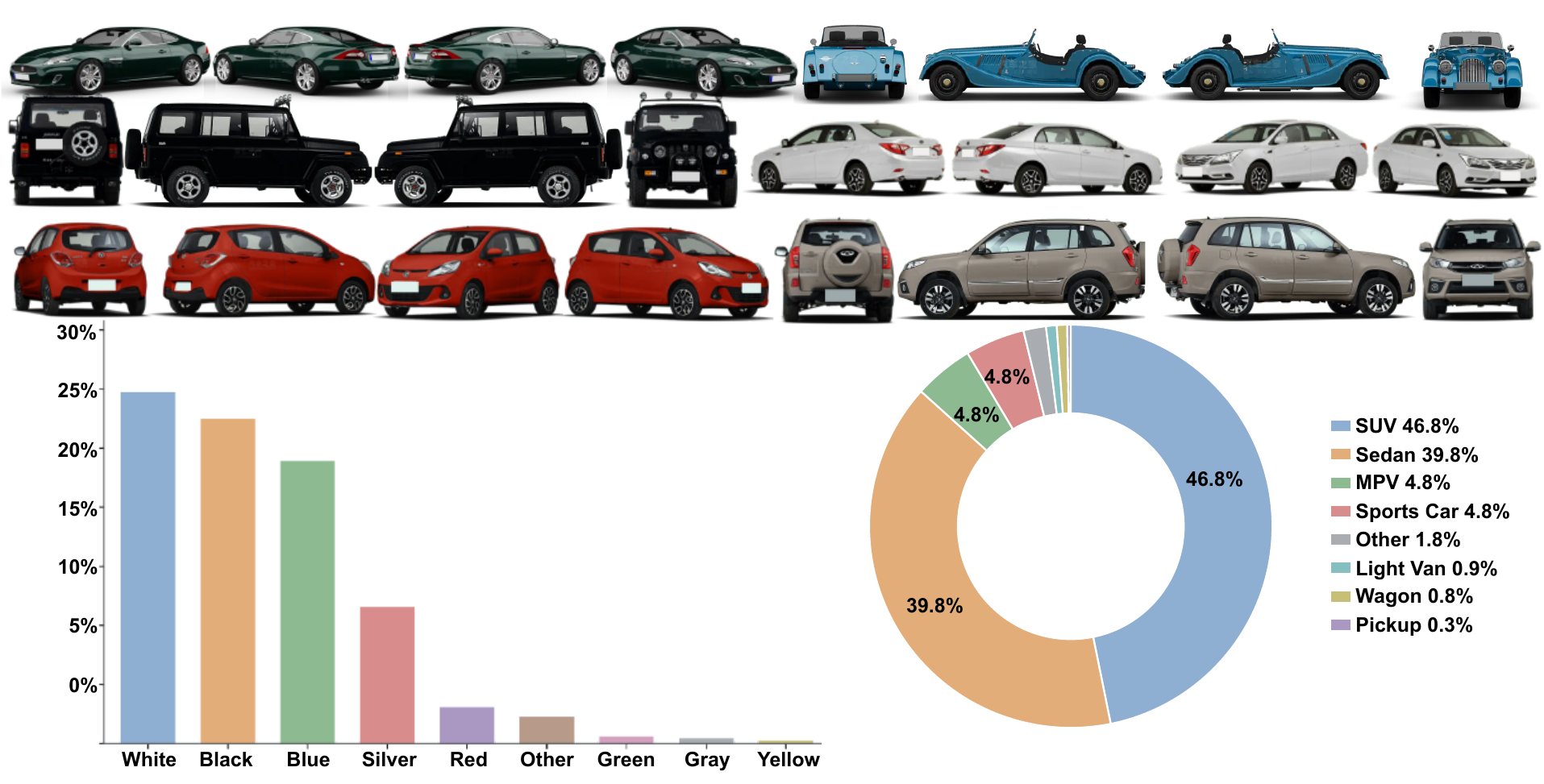}
    \vspace{-6mm}
    \caption{\textbf{RECAR dataset.} Examples of representative multi-view RGBA vehicle images, 
    together with a bar chart illustrating the vehicle color distribution and a pie chart illustrating the vehicle type distribution.}
    \label{fig:RECAR_dataset}
    \vspace{-4mm}
\end{figure}

\noindent\textbf{RECAR: Real-Vehicle Multi-View Dataset. }
\label{sec:RECAR}
SRN-Car~\cite{srn} and Objaverse-XL-Car~\cite{objaverse} provide synthetic
assets whose appearance and geometric detail can differ from real vehicles.
MVMC~\cite{mvmc}, 3DRealCar~\cite{3drealcar}, and MAD-CARS~\cite{madrive}
collect real-world observations, but include backgrounds and lack standardized
viewpoint sampling, as Table~\ref{tab:vehicle_datasets} shows.
RECAR provides 20K real vehicles represented by 600K background-free RGBA
images at $1200\times800$ resolution. We organize the images by vehicle
identity into full $360^\circ$ sequences with fixed elevation and regular
azimuth sampling. Figure~\ref{fig:RECAR_dataset} shows representative views, the vehicle color distribution, and the vehicle type 
distribution. These background-free,
regularly sampled views provide consistent multi-view inputs for pseudo-target
construction.

\noindent\textbf{Pseudo-target construction. }
For each RECAR vehicle \(k\), we feed its front, rear, left, and right views
\(\mathcal{I}_k^{T}\) to the original TRELLIS model to construct pseudo-targets.
Using TRELLIS's default stochastic multi-image conditioning, we obtain
\begin{equation}
    (\mathbf{S}_k^{T},\mathbf{P}_k^{T},\mathbf{Z}_k^{T})
    =
    \mathcal{T}_{\mathrm{mv}}(\mathcal{I}_k^{T}),
    \qquad
    \mathcal{A}_k^{T}
    =
    \mathcal{D}_{\mathrm{GS}}
    (\mathbf{Z}_k^{T};\mathbf{P}_k^{T}),
\end{equation}
where $\mathcal{T}_{\mathrm{mv}}$ denotes TRELLIS generation with multi-image
conditioning and $\mathcal{D}_{\mathrm{GS}}$ is its 3D Gaussian decoder.
Here, $\mathbf{S}_k^{T}$ denotes the sparse structure (SS) latent,
$\mathbf{P}_k^{T}$ denotes the decoded active voxels,
$\mathbf{Z}_k^{T}$ is the structured latent (SLAT) defined over these voxels,
and $\mathcal{A}_k^{T}$ is the decoded 3D Gaussian asset.

We use the pretrained Improved Aesthetic Predictor~\cite{aesthetic_predictor}
to assess the visual quality of each generated asset.
We retain generated assets with aesthetic scores above 4.5, manually screen
their vehicle geometry, and cache the sparse structure latents and SLAT representations
of selected assets as training pseudo-targets.

\begin{table}[!t]
\centering
\caption{\textbf{Comparison of vehicle-centric datasets.}}
\vspace{-2mm}
\label{tab:vehicle_datasets}
\begingroup
\definecolor{datasetYes}{RGB}{30,125,65}
\definecolor{datasetNo}{RGB}{185,45,45}
\newcommand{\datasetcheck}{\textcolor{datasetYes}{$\checkmark$}}
\newcommand{\datasetcross}{\textcolor{datasetNo}{$\times$}}
\scriptsize
\setlength{\tabcolsep}{2.0pt}
\renewcommand{\arraystretch}{1.10}

\resizebox{\columnwidth}{!}{%
\begin{tabular}{lcccccc}
\toprule
\textbf{Dataset}
& \textbf{Vehicles}
& \textbf{Type}
& \textbf{Views}
& \textbf{Resolution}
& \textbf{BG-free}
& \textbf{Regular} \\
\midrule
SRN-Car~\cite{srn}
& 2.1K & Synthetic & 250 & $128{\times}128$ & \datasetcheck & \datasetcheck \\

Objaverse-XL-Car~\cite{objaverse}
& 11.4K & Synthetic & -- & -- & \datasetcheck & \datasetcheck \\

MVMC~\cite{mvmc}
& 576 & Real & 10 & $600{\times}450$ & \datasetcross & \datasetcross \\

3DRealCar~\cite{3drealcar}
& 2.5K & Real & 200 & $1920{\times}1440$ & \datasetcross & \datasetcross \\

MAD-CARS~\cite{madrive}
& 70K & Real & 85 & $1920{\times}1080$ & \datasetcross & \datasetcross \\

\midrule
\textbf{RECAR}
& 20K & Real & 30 & $1200{\times}800$ & \datasetcheck & \datasetcheck \\
\bottomrule
\end{tabular}%
}

\vspace{2pt}
\parbox{\columnwidth}{\scriptsize \textit{Note:}
\datasetcheck / \datasetcross: yes / no. --: not applicable. Views denotes the average number of views or 
video frames available per vehicle. 
BG-free indicates whether background-free images are provided or can be rendered from the supplied assets.
Regular denotes viewpoint sampling according to a predefined pattern.
}
\vspace{-4mm}
\endgroup
\end{table}

\noindent\textbf{Stage I: vehicle-domain adaptation. }
We fine-tune TRELLIS's sparse structure and SLAT flow transformers on
vehicle geometry and appearance. For each retained RECAR vehicle \(k\), we
sample a single clean RGBA image \(I_{k,n}\) as the conditioning observation
and optimize against the cached pseudo-targets:
\begin{equation}
    \theta_{\mathrm{veh}}
    =
    \operatorname*{arg\,min}_{\theta}
    \mathbb{E}_{k,n}
    \left[
        \mathcal{L}_{\mathrm{adapt}}
        \left(
            I_{k,n};
            \mathbf{S}_{k}^{T},
            \mathbf{P}_{k}^{T},
            \mathbf{Z}_{k}^{T}
        \right)
    \right],
\end{equation}
where $\mathcal{L}_{\mathrm{adapt}}$ follows the original flow-matching
objective of TRELLIS~\cite{trellis}.

\noindent\textbf{Stage II: driving-log-robust adaptation. }
The clean RGBA images used in Stage I differ from actor crops extracted
from real driving logs. To adapt to these inputs, we retain the clean RGBA observation \(I_{k,n}\) with
probability \(\rho\); otherwise, we apply synthetic perturbations that mimic
real driving-log imagery and imperfect foreground masks.
We denote the sampled perturbation by \(\mathcal{D}\) and the resulting
conditioning image by \(I_{k,n}^{\mathrm{cond}}\).
Starting from $\theta_{\mathrm{veh}}$, we optimize
\begin{equation}
    \theta_{\mathrm{rob}}
    =
    \operatorname*{arg\,min}_{\theta}
    \mathbb{E}_{k,n,\mathcal{D}}
    \left[
        \mathcal{L}_{\mathrm{adapt}}
        \left(
            I_{k,n}^{\mathrm{cond}};
            \mathbf{S}_{k}^{T},
            \mathbf{P}_{k}^{T},
            \mathbf{Z}_{k}^{T}
        \right)
    \right].
\end{equation}
We perturb only the conditioning observation and keep the cached pseudo-targets
unchanged, training the model to recover vehicle geometry and appearance
from real inputs.

\noindent\textbf{Inference. }
At inference, we localize the target actor using Grounding
DINO~\cite{groundingdino} and obtain an alpha mask using box-prompted
SAM~\cite{sam}. Given the single-view RGBA crop, the adapted model first generates
the sparse structure and then the associated latent features,
forming a SLAT representation that is decoded into a view-complete
3D Gaussian actor $\widehat{\mathcal{G}}_{i}$.

\subsection{Actor Registration and Insertion}
\label{sec:scene_simulation}
Actor replacement requires geometric alignment with the logged vehicle while
preserving the scene's calibration and pose conventions. We register the
generated asset in the actor-local frame so that it can use the target
vehicle's local-to-world pose mapping.

\noindent\textbf{LiDAR-guided actor registration. }
Logged 3D boxes provide an initial scale and orientation, while LiDAR supplies
geometric evidence for refining the alignment. We convert the TRELLIS
coordinate axes, align the generated actor's yaw, and scale its bounding box
to the logged vehicle box. To combine observations of the moving vehicle,
we aggregate LiDAR points within its logged boxes across visible frames in
the actor-local frame. Constrained Iterative Closest Point (ICP)~\cite{ICP}
aligns the generated Gaussian centers with this point cloud by estimating an
isotropic scale, a yaw offset, and a 3D translation.

\noindent\textbf{Actor insertion. }
Street Gaussians~\cite{streetgaussians} separates the world-frame background
$\mathcal{G}_{\mathrm{bg}}$ from actor-local Gaussian models
$\{\mathcal{G}_{j}^{\mathrm{log}}\}$, allowing us to replace the target's
geometry while retaining its pose mapping. We replace
$\mathcal{G}_{i}^{\mathrm{log}}$ with the registered view-complete actor
$\widehat{\mathcal{G}}_{i}$ and retain $\mathbf{T}_{i,\mathrm{log}}^{t}$,
which maps the actor-local frame to the world frame.
To approximate the actor's contact shadow, we add a flattened black Gaussian
primitive on its canonical ground plane. The actor and shadow share the
registration transform, which updates their Gaussian centers, covariances,
and normals, keeping them aligned during subsequent pose updates.

\subsection{Closed-Loop Simulation}
\label{sec:closed_loop_simulation}
RECAST supports state-conditioned and image-conditioned ego planning with
editable actor motion. At simulation cycle $n$, the current actor pose defines
its placement in the Gaussian scene $\mathcal{G}_{\mathrm{sim}}^{n}$.

\noindent\textbf{Ego controller. }
The ego planner receives
\begin{equation}
    \mathbf{o}_{m}^{n}
    =
    \begin{cases}
        (\mathcal{H}_{e}^{n},\mathcal{S}_{\mathrm{actor}}^{n},
         \mathcal{M}_{\mathrm{HD}},\mathbf{q}_{\mathrm{state}}),
        & m=\mathrm{state}, \\
        (\mathcal{R}(\mathcal{G}_{\mathrm{sim}}^{n},\mathbf{C}^{n}),
         \mathbf{s}_{e}^{n}),
        & m=\mathrm{image},
    \end{cases}
\end{equation}
where $\mathbf{o}_{m}^{n}$ is the planner input and $m$ denotes its modality.
For state-conditioned planning, $\mathcal{H}_{e}^{n}$ is the ego history,
$\mathcal{S}_{\mathrm{actor}}^{n}$ the current actor states,
$\mathcal{M}_{\mathrm{HD}}$ the HD map, and $\mathbf{q}_{\mathrm{state}}$
the navigation input. For image-conditioned planning, $\mathcal{R}$ renders
camera images from the scene at camera poses $\mathbf{C}^{n}$, and
$\mathbf{s}_{e}^{n}$ is the ego state. Subscripts $e$ and $a$ denote the ego
vehicle and target actor, respectively.

The ego planner predicts a trajectory, and the ego controller executes its first
$K$ waypoints:
\begin{align}
    \widehat{\boldsymbol{\tau}}_{e}^{n}
    &= \pi_e^m(\mathbf{o}_{m}^{n}), \\
    (\mathbf{s}_{e}^{n+1},\mathbf{C}^{n+1})
    &= \mathcal{F}_{e}
    (\mathbf{s}_{e}^{n},\widehat{\boldsymbol{\tau}}_{e,1:K}^{n};\Delta t),
\end{align}
where $\pi_e^m$ is the ego planner for modality $m$,
$\widehat{\boldsymbol{\tau}}_{e}^{n}$ is its predicted trajectory, and
$\widehat{\boldsymbol{\tau}}_{e,1:K}^{n}$ contains the first $K$ waypoints.
The update function $\mathcal{F}_{e}$ advances the ego state and camera poses
over the execution time interval $\Delta t$.

\noindent\textbf{Actor controller. }
The specified actor motion defines a path and heading, while a speed
controller based on the Intelligent Driver Model (IDM)~\cite{IDM} tracks
the desired speed along the path. The simulator updates the actor over
the same time interval:
\begin{equation}
    \mathbf{s}_{a}^{n+1}
    = \mathcal{F}_{a}
    (\mathbf{s}_{a}^{n},\pi_a,\mathcal{P}_{a};\Delta t),
\end{equation}
where $\mathbf{s}_{a}^{n}$ contains the actor's position, speed, and heading,
$\pi_a$ is its speed controller configured with the desired speed,
$\mathcal{P}_{a}$ specifies its path and heading, and $\mathcal{F}_{a}$ is the
actor state update function. 

The updated ego and actor states determine the
next planner input, using state information for state-conditioned planning
and rendered images with the ego state for image-conditioned planning.

\section{Experiments}
\label{sec:experiments}
We mainly investigate the following three research questions based on the proposed RECAST:

\begingroup
\renewcommand{\theenumi}{Q\arabic{enumi}}
\renewcommand{\labelenumi}{\textbf{\theenumi:}}
\begin{enumerate}[\settowidth{\labelwidth}{\textbf{Q3:}}\setlength{\labelsep}{0.5em}\setlength{\itemsep}{0.5em}]
\item\label{rq:q1}
Can RECAST support closed-loop render--plan--update simulation with a
view-complete actor generated from a logged image?

\item\label{rq:q2}
Does RECAST improve scene-level rendering fidelity under controlled
ego--actor interactions compared with native actors and alternative
actor-generation methods using the same reconstructed background?

\item\label{rq:q3}
Do vehicle-domain adaptation on RECAR and driving-log-robust adaptation
improve single-image vehicle generation?

\end{enumerate}
\endgroup

\subsection{Experimental Setup}
\label{sec:experimental_setup}

\noindent\textbf{Datasets. }
For closed-loop simulation and scene-level evaluation, we construct a
controlled-interaction benchmark from nine Waymo Open Dataset~\cite{waymo}
scenes (002, 011, 034, 046, 137, 152, 161, 162, and 177). In each scene,
we select one surrounding vehicle as the target actor and remove all other
dynamic actors. The logged sequence provides reference observations, while
controlled target motion exposes the actor from different viewpoints relative
to the ego vehicle.

After the quality filtering described in Sec.~\ref{sec:RECAR}, RECAR retains 12,896 
vehicle identities, which are split into 12,248 training instances and 648 
held-out test instances for actor-level training and evaluation.

\noindent\textbf{Implementation details. }
For scene-level experiments, RECAST uses Street Gaussians as its scene-reconstruction backbone.
Each scene is optimized for 30,000 iterations with a learning rate of \(1\times10^{-5}\) on a single NVIDIA A40 GPU. 
We render each rollout at \(960\times640\) resolution using the three forward-facing Waymo cameras (left-front, front, 
and right-front). Simulation and rendering use a simulation time step of \(0.1\,\mathrm{s}\), and each rollout contains 160 frames (\(16\,\mathrm{s}\)).

For vehicle-domain adaptation, we initialize the sparse structure and SLAT 
flow-matching Transformers from the official TRELLIS models and optimize both with AdamW (\(\mathrm{lr}=1\times10^{-5}\)) on 
two NVIDIA A40 GPUs, using an effective batch size of 2, EMA decay \(0.9999\), and classifier-free dropout \(p_{\mathrm{uncond}}=0.1\). 
We use the  sparse structure
checkpoint at 10,000 steps and the SLAT checkpoint at 16,000 steps for subsequent adaptation.

For driving-log-robust adaptation, we continue training from these checkpoints
with a learning rate of \(2\times10^{-6}\) for 12,000 steps. Clean conditioning is
retained with probability \(\rho=0.2\); otherwise, we sample image corruptions that mimic driving observations, including down--up sampling, blur, compression, low-light
corruption, fog, occlusion, cropping, and foreground-mask perturbations.

\subsection[Q1: Closed-Loop Simulation with Ego Planners]{\textbf{\hyperref[rq:q1]{Q1}}: Closed-Loop Simulation with Ego Planners}
\textbf{\hyperref[rq:q1]{Q1}} is the central question motivating RECAST.
We evaluate how actor rendering affects closed-loop planner evaluation
with the image-conditioned planner GTRS-Dense~\cite{gtrs}.

We use three actor motions on Waymo: cutting in front of the ego vehicle, approaching the ego vehicle from the opposite direction, and remaining stationary while facing the ego vehicle.
Each scenario uses
two configurations with different initial distances and motion parameters.
We compare actor representations using Street Gaussians and OmniRe backgrounds,
removing all non-target dynamic actors and using identical target-actor states across methods. At each planning cycle, GTRS-Dense receives the current
multi-camera rendering and predicts 40 waypoints at 0.1 s intervals,
covering a 4 s horizon. The ego vehicle executes the first five waypoints,
and the updated ego and target actor states determine the next rendering.
Each rollout consists of 32 planning cycles and spans 16 s.

We assess driving safety using no-collision rate (NC), drivable-area
compliance (DAC), and time-to-collision (TTC), and report route completion
($R_c$) and comfort (COM) to evaluate progress and motion smoothness.
NC is the percentage of collision-free rollouts. We detect collisions by checking
bird's-eye-view overlap between the ego vehicle's and target actor's 3D bounding
boxes. All methods use boxes from the corresponding log frames, transformed
to the vehicles' current simulation poses. DAC is the percentage of rollouts in which 
the ego vehicle remains within the map-defined drivable area throughout the rollout. TTC is
the minimum predicted collision time across 32 planning cycles, capped
at 4 s and averaged over rollouts. $R_c$ normalizes distance traveled
before collision by the remaining reference-route length from beginning to end.
COM checks each 40-waypoint prediction against limits on longitudinal
acceleration and jerk, yaw rate and acceleration, and lateral acceleration.

We compute 95\% confidence intervals for paired metric differences 
(RECAST minus native Street Gaussians actors) using 200,000 scene-level bootstrap samples.
We resample the nine scenes with replacement, retaining all six matched rollouts 
per scene, and take the 2.5th and 97.5th percentiles of the mean differences.

Table~\ref{tab:closed_loop_rollout} shows different closed-loop outcomes
for the same GTRS-Dense planner across actor representations. Comparisons
within each background use the same scenes, interactions, initial states,
target controller, and reference routes. With the Street Gaussians background
fixed, the planner achieves an NC rate of 22.2\% with native actors and 63.0\%
with RECAST actors, a paired difference of 40.7 percentage points
(95\% CI: [25.93, 55.56]). TTC differs by 1.352 s
(95\% CI: [0.628, 2.046]) and COM by 4.7 percentage points
(95\% CI: [2.1, 7.5]), both higher with RECAST actors. Route completion is
16.82\% and 17.57\%, respectively.

\begin{table}[!t]
\centering
\caption{\textbf{Image-conditioned closed-loop evaluation with GTRS-Dense~\cite{gtrs}.}}
\vspace{-2mm}
\label{tab:closed_loop_rollout}
\begingroup
\scriptsize
\setlength{\tabcolsep}{1.5pt}
\setlength{\fboxsep}{1pt}
\renewcommand{\arraystretch}{1.10}
\resizebox{\columnwidth}{!}{%
\begin{tabular}{@{}lccccc@{}}
\toprule
\textbf{Method}
& NC (\%) $\uparrow$
&DAC (\%) $\uparrow$
& TTC (s) $\uparrow$
& $R_c$ (\%) $\uparrow$
& COM (\%) $\uparrow$ \\
\midrule
StreetGS~\cite{streetgaussians}
& \begin{tabular}[t]{@{}c@{}}12/54\\[-3pt]{\tiny(22.2\%)}\end{tabular} &\begin{tabular}[t]{@{}c@{}}53/54\\[-3pt]{\tiny(98.1\%)}\end{tabular} & 0.798 & 16.82& 91.20\\
\quad w/ TRELLIS~\cite{trellis}
& \begin{tabular}[t]{@{}c@{}}33/54\\[-3pt]{\tiny(61.1\%)}\end{tabular}&\begin{tabular}[t]{@{}c@{}}53/54\\[-3pt]{\tiny(98.1\%)}\end{tabular} & 2.115& 16.81& 94.68\\
\quad w/ AH~\cite{asset_harvester}
& \begin{tabular}[t]{@{}c@{}}30/54\\[-3pt]{\tiny(55.6\%)}\end{tabular} &\colorbox{bestcolor}{\begin{tabular}[t]{@{}c@{}}54/54\\[-3pt]{\tiny(100\%)}\end{tabular}} & 2.070 & 17.11& 94.91\\

OmniRe~\cite{omnire}
& \begin{tabular}[t]{@{}c@{}}19/54\\[-3pt]{\tiny(35.2\%)}\end{tabular} &\begin{tabular}[t]{@{}c@{}}53/54\\[-3pt]{\tiny(98.1\%)}\end{tabular} & 1.178 & 16.67& 89.12\\
\quad w/ TRELLIS~\cite{trellis}
& \begin{tabular}[t]{@{}c@{}}32/54\\[-3pt]{\tiny(59.3\%)}\end{tabular} &\begin{tabular}[t]{@{}c@{}}52/54\\[-3pt]{\tiny(96.3\%)}\end{tabular} & 2.139 & 16.72& 93.87\\
\quad w/ AH~\cite{asset_harvester}
& \begin{tabular}[t]{@{}c@{}}33/54\\[-3pt]{\tiny(61.1\%)}\end{tabular} &\begin{tabular}[t]{@{}c@{}}53/54\\[-3pt]{\tiny(98.1\%)}\end{tabular} & 2.119 & 16.44& 94.44\\
\textbf{Ours}
& \colorbox{bestcolor}{\begin{tabular}[t]{@{}c@{}}34/54\\[-3pt]{\tiny(63.0\%)}\end{tabular}}&\colorbox{bestcolor}{\begin{tabular}[t]{@{}c@{}}54/54\\[-3pt]{\tiny(100\%)}\end{tabular}} & \colorbox{bestcolor}{2.150} & \colorbox{bestcolor}{17.57}& \colorbox{bestcolor}{95.89}\\
\bottomrule
\end{tabular}%
}

\vspace{2pt}
\parbox{\columnwidth}{\scriptsize
\textit{Note:} Each method is evaluated on 54 deterministic 
rollouts across nine scenes, three motions, and two actor distance--speed 
settings. We report no-collision rate (NC), drivable-area compliance (DAC), 
mean minimum predicted time-to-collision (TTC), route completion ($R_c$), and driving 
comfort (COM). w/ denotes replacement of the native target actor with a generated actor. AH denotes Asset Harvester~\cite{asset_harvester}. We highlight the \colorbox{bestcolor}{best} results.
}
\vspace{-4mm}
\endgroup
\end{table}

The differing outcomes with a fixed planner show that closed-loop
evaluation depends on actor rendering. For each scene, all methods 
use the target actor's bounding boxes
from the corresponding log frames, updated to its simulated pose. 
Within each background, actor replacement changes the visual 
observations on which the planner bases its decisions.

Beyond logged trajectories, incomplete reconstructions expose missing
or distorted vehicle surfaces. These artifacts can alter a planning
decision, changing the ego pose and subsequent observations. This
feedback makes rendering errors a possible source of closed-loop
failures, complicating their attribution to the planner. RECAST
addresses this source of error by supplying a view-complete actor
as viewpoints change, supporting planner-in-the-loop evaluation
under controlled interactions beyond log replay.

\subsection[Q2: Scene-Level Rendering Quality]{\textbf{\hyperref[rq:q2]{Q2}}: Scene-Level Rendering Quality}
\label{sec:scene_evaluation}
\begin{figure*}[!t]
    \centering
    \includegraphics[width=1\linewidth]{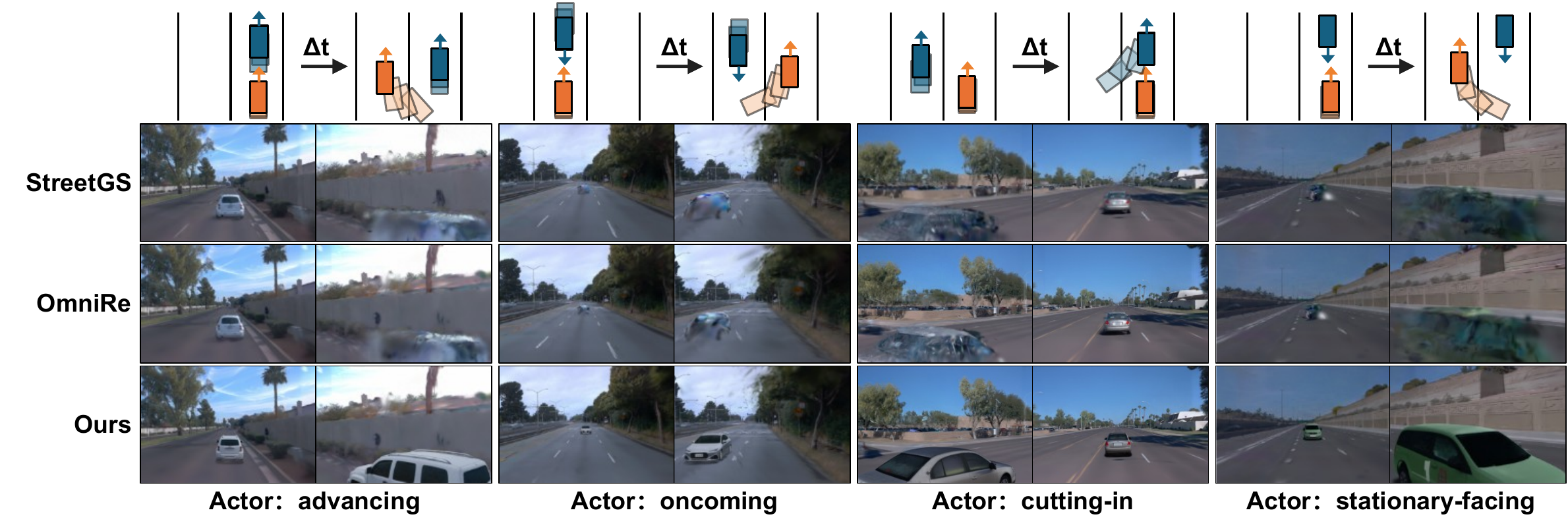}
    \vspace{-6mm}
    \caption{
    \textbf{Scene-level qualitative comparison under controlled interactions on Waymo.}
We evaluate four target-actor motion settings. All methods share the same
ego--actor trajectories within each scenario.\quad\mbox{\protect\textcolor{egoColor}{\protect\rule{0.8em}{0.8em}}~Ego\quad
\protect\textcolor{actorColor}{\protect\rule{0.8em}{0.8em}}~Actor}}
    \label{fig:scene_qualitative}
    \vspace{-4mm}
\end{figure*}

To address \textbf{\hyperref[rq:q2]{Q2}}, we compare RECAST
with Street Gaussians and OmniRe using their native actors
and actors generated by TRELLIS or Asset Harvester in
Table~\ref{tab:overtaking_main} and Fig.~\ref{fig:scene_qualitative}.

\begin{table}[!htb]
\centering
\caption{\textbf{Scene-level quantitative comparison under controlled interactions on Waymo.}
}
\vspace{-2mm}
\label{tab:overtaking_main}

\begingroup
\scriptsize
\setlength{\tabcolsep}{1.5pt}
\setlength{\fboxsep}{1pt}
\renewcommand{\arraystretch}{1.10}

\begin{tabular*}{\columnwidth}{@{\extracolsep{\fill}}lccc@{}}
\toprule
\multicolumn{4}{c}{\textbf{Scene quality}} \\
\midrule
\textbf{Method} & $\mathrm{FD}_{\mathrm{incep}}\downarrow$ & $\mathrm{CLIP}_{\mathrm{margin}}\uparrow$ & $\mathrm{VQA}_{\mathrm{margin}}\uparrow$ \\
\midrule
StreetGS~\cite{streetgaussians} & 129.35 & 0.14 & -294.91 \\
\quad w/ TRELLIS~\cite{trellis} & 112.84 & 2.86 & -172.36 \\
\quad w/ AH~\cite{asset_harvester} & \colorbox{bestcolor}{110.00} & 1.95 & \colorbox{secondcolor}{-156.42} \\
OmniRe~\cite{omnire} & 138.52 & 0.04 & -314.18 \\
\quad w/ TRELLIS~\cite{trellis} & 121.47 & \colorbox{secondcolor}{2.93} & -216.82 \\
\quad w/ AH~\cite{asset_harvester} & 119.10 & 1.96 & -197.11 \\
\textbf{Ours} & \colorbox{secondcolor}{112.10} & \colorbox{bestcolor}{3.47} & \colorbox{bestcolor}{-131.04} \\
\bottomrule
\end{tabular*}

\par\vspace{4pt}
\begin{tabular*}{\columnwidth}{@{\extracolsep{\fill}}lcccc@{}}
\toprule
\multicolumn{5}{c}{\textbf{Actor-crop quality}} \\
\midrule
\textbf{Method} & $\mathrm{FD}_{\mathrm{incep}}\downarrow$ & $\mathrm{CLIP}_{\mathrm{img}}\uparrow$ & $\mathrm{CLIP}_{\mathrm{veh}}\uparrow$ & $\mathrm{LPIPS}\downarrow$ \\
\midrule
StreetGS~\cite{streetgaussians} & 193.56 & 65.43 & 20.62 & 0.6920 \\
\quad w/ TRELLIS~\cite{trellis} & 126.14 & 69.54 & 21.28 & \colorbox{secondcolor}{0.6809} \\
\quad w/ AH~\cite{asset_harvester} & \colorbox{bestcolor}{124.61} & 70.17 & 21.03 & 0.6818 \\
OmniRe~\cite{omnire} & 197.41 & 65.72 & 20.87 & 0.6931 \\
\quad w/ TRELLIS~\cite{trellis} & 133.04 & 69.99 & \colorbox{secondcolor}{21.32} & 0.6824 \\
\quad w/ AH~\cite{asset_harvester} & 129.11 & \colorbox{secondcolor}{70.23} & 21.11 & 0.6823 \\
\textbf{Ours} & \colorbox{secondcolor}{124.74} & \colorbox{bestcolor}{70.67} & \colorbox{bestcolor}{21.43} & \colorbox{bestcolor}{0.6800} \\
\bottomrule
\end{tabular*}

\vspace{2pt}
\parbox{\columnwidth}{\scriptsize
\textit{Note:} $\mathrm{CLIP}_{\mathrm{veh}}$ uses the fixed prompt ``vehicle.'' Margin denotes the 
mean positive-prompt score minus the mean negative-prompt score. Margin metrics are 
reported $\times1000$. $\mathrm{CLIP}_{\mathrm{img}}$ and $\mathrm{CLIP}_{\mathrm{veh}}$ are 
reported $\times100$.
w/ denotes replacement of the native target actor with a generated actor.  AH denotes Asset Harvester~\cite{asset_harvester}. 
We highlight the \colorbox{bestcolor}{best} and \colorbox{secondcolor}{second-best} results.
}
\vspace{-4mm}
\endgroup
\end{table}

For each Waymo scene, we retain the three actor motion types from the closed-loop evaluation and add advancing motion, 
during which the ego vehicle overtakes the actor.
PDM-Closed~\cite{pdmclosed} generates one
map-conditioned ego trajectory for each scene--configuration pair. We cache
the resulting ego and target actor poses and reuse them across renderers.
All methods therefore share the same initial states, trajectories, camera
calibration, reference observations, and evaluated frames, isolating the
effect of the actor representation within each background.

We use $\mathrm{FD}_{\mathrm{incep}}$~\cite{fid} to assess scene and actor-crop realism
against real Waymo observations containing the target actor.
$\mathrm{CLIP}_{\mathrm{img}}$~\cite{clip} and LPIPS~\cite{lpips} measure similarity
to reference vehicle crops without matched viewpoints, while
$\mathrm{CLIP}_{\mathrm{veh}}$ measures alignment with the prompt ``vehicle.''
For each rendered image, $\mathrm{CLIP}_{\mathrm{margin}}$ and
$\mathrm{VQA}_{\mathrm{margin}}$ are defined as the mean positive-prompt score
minus the mean negative-prompt score, computed using CLIP and
VQAScore~\cite{vqascore}, respectively. Higher margins indicate stronger alignment with descriptions
of realistic, complete vehicles.

All prompts begin with ``A driving-camera image with''.
Positive endings are ``realistic vehicles,''
``complete and recognizable vehicles,'' and
``vehicles that have normal shapes and intact geometry.''
Negative endings are ``unrealistic vehicles,''
``incomplete or unrecognizable vehicles,'' and
``vehicles that have distorted shapes or fragmented geometry.''
We use the same prompts for evaluation of $\mathrm{CLIP}_{\mathrm{margin}}$ and
$\mathrm{VQA}_{\mathrm{margin}}$.

With the Street Gaussians background fixed, RECAST achieves the best CLIP and
VQA margins, $\mathrm{CLIP}_{\mathrm{img}}$, $\mathrm{CLIP}_{\mathrm{veh}}$, and LPIPS among the
compared actor representations. Asset Harvester achieves the lowest scene
and target-crop $\mathrm{FD}_{\mathrm{incep}}$ (110.00 and 124.61), followed by RECAST (112.10 and 124.74). Figure~\ref{fig:scene_qualitative} illustrates
the improved vehicle appearance under controlled motion. Relative to native Street Gaussians actors, RECAST reduces target-crop $\mathrm{FD}_{\mathrm{incep}}$
by 35.6\% and scene $\mathrm{FD}_{\mathrm{incep}}$ by 13.3\%.

Fixed trajectories isolate rendering differences from the planner's
response. Scene scores assess the actor within the shared scene,
while crop scores focus on the replaced vehicle. The different metric
rankings reflect distinct evaluation targets: $\mathrm{FD}_{\mathrm{incep}}$ compares image
distributions, whereas reference-crop metrics compare individual vehicle
appearances. Asset Harvester's lower $\mathrm{FD}_{\mathrm{incep}}$ therefore does not establish
a closer match to the target vehicle.

Viewpoint differences limit the reference-crop scores to evidence of
appearance similarity, while the qualitative examples show coherent
vehicle shape across viewpoint changes. RECAST achieves $\mathrm{FD}_{\mathrm{incep}}$ close to
Asset Harvester's and better reference-crop scores, supporting its ability
to render view-complete vehicles beyond logged viewpoints while retaining
appearance cues from the input.

\subsection[Q3: Actor-Level Generation Quality]{\textbf{\hyperref[rq:q3]{Q3}}: Actor-Level Generation Quality}
\begin{figure*}[!t]
    \centering
    \includegraphics[width=1\linewidth]{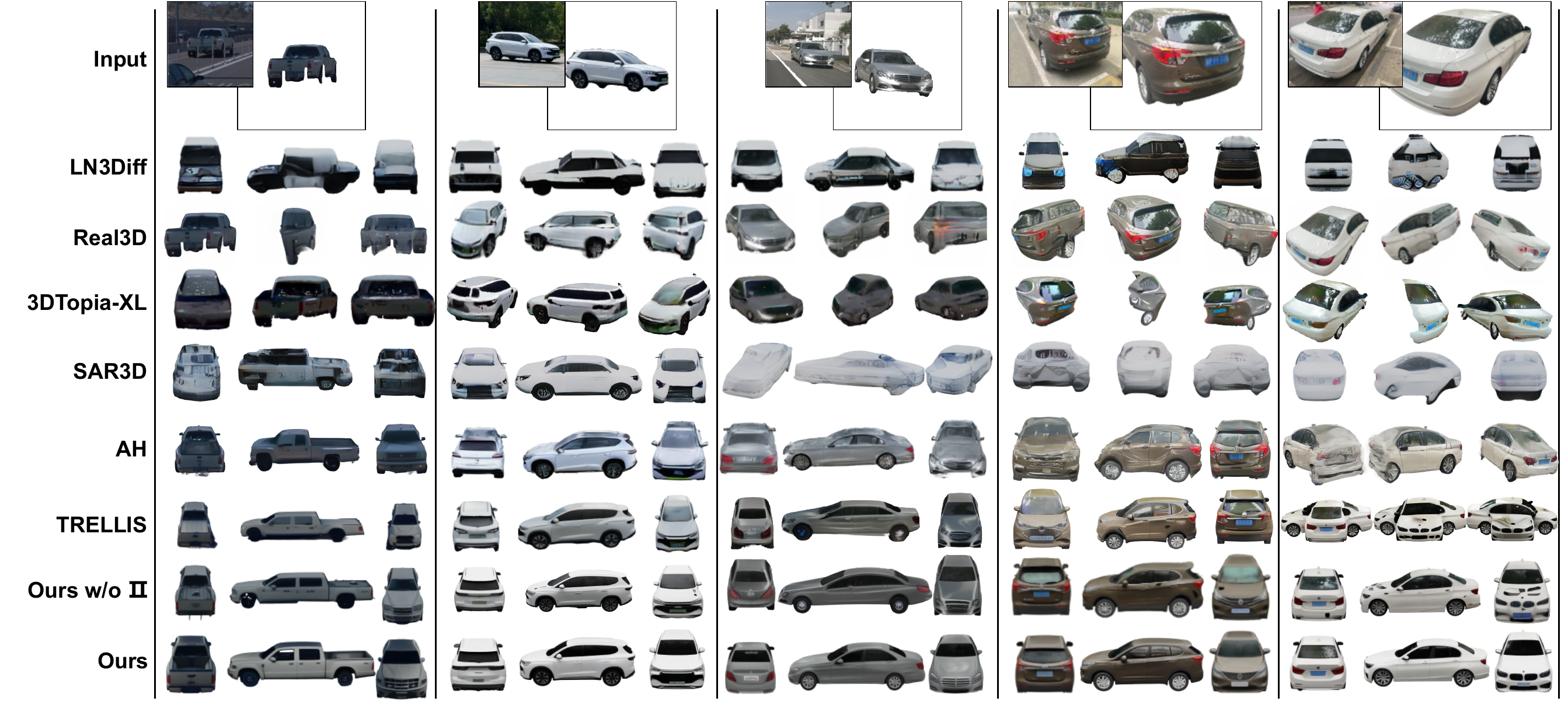}
    \caption{\textbf{Actor-level qualitative comparison of single-image vehicle generation.}
    The first three examples are from driving-log crops, while the last two examples 
    are from 3DRealCar~\cite{3drealcar}. AH denotes Asset Harvester~\cite{asset_harvester}, and Ours w/o II denotes our model without driving-log-robust adaptation.
    }
    \label{fig:actor_qualitative}
    \vspace{-4mm}
\end{figure*}

To address \textbf{\hyperref[rq:q3]{Q3}}, we compare single-image vehicle generation
methods and assess the two adaptation stages in Fig.~\ref{fig:actor_qualitative} and Table~\ref{tab:vehicle_generation}.

Each method receives the same single RGBA observation and renders four target
views excluded from the conditioning input. Reference images are the original
photographs of the corresponding RECAR vehicles, provided in RGBA format.
The target cameras use a $40^\circ$
field of view, a $15^\circ$ pitch, and azimuths of $0^\circ$, $90^\circ$,
$180^\circ$, and $270^\circ$, with identical parameters for rendered and
reference views. We use PSNR, SSIM~\cite{ssim}, LPIPS, and $\mathrm{CLIP}_{\mathrm{img}}$
to assess visual fidelity and perceptual similarity, and
$\mathrm{FD}_{\mathrm{incep}}$ and $\mathrm{FD}_{\mathrm{dinov2}}$~\cite{fid,dinov2} to evaluate
the overall quality of generated vehicles. 
RECAST outperforms the evaluated baselines on all six metrics
in Table~\ref{tab:vehicle_generation}. Figure~\ref{fig:actor_qualitative}
shows more complete vehicle geometry and closer appearance matches to the input observations. 
The gains in both distributional and reference-based metrics
indicate that RECAST improves vehicle realism while preserving the target
vehicle's appearance in views absent from the conditioning image.

\begin{table}[!t]
    \centering
    \caption{
    \textbf{Actor-level quantitative generation results on held-out RECAR vehicles.}
    }
    \vspace{-2mm}
    \label{tab:vehicle_generation}

    \begingroup
    \scriptsize
    \setlength{\tabcolsep}{1.5pt}
\setlength{\fboxsep}{1pt}
    \renewcommand{\arraystretch}{1.10}

    \resizebox{\columnwidth}{!}{%
\begin{tabular}{
        @{}
        lcccccc
        @{}
    }
        \toprule
        Method
        & PSNR$\uparrow$
        & SSIM$\uparrow$
        & LPIPS$\downarrow$
        & $\mathrm{CLIP}_{\mathrm{img}}\uparrow$
        & $\mathrm{FD}_{\mathrm{incep}}\downarrow$
        & $\mathrm{FD}_{\mathrm{dinov2}}\downarrow$
        \\
        \midrule

        Real3D~\cite{real3d}
        & 13.19 & 0.5691 & 0.3223
        & 68.17 & 91.196 & 1282.01 \\

        LN3Diff~\cite{ln3diff}
        & 18.95 & 0.8843 & 0.1105
        & 63.03 & 93.288 & 1455.05 \\

        3DTopia-XL~\cite{3dtopia-xl}
        & 15.54 & 0.8085 & 0.1817
        & 59.90 & 166.294 & 1299.59 \\

        SAR3D~\cite{sar3d}
        & 15.39 & 0.8231 & 0.1588
        & 66.55 & 64.042 & 746.99 \\

        AH~\cite{asset_harvester}
        & 19.22 & 0.8792 & 0.1044
        & 78.29
        & 38.368 & 183.08 \\

        TRELLIS~\cite{trellis}
        & \colorbox{secondcolor}{21.02} & 0.9077 & 0.0603
        & 78.22 & 9.788 & 43.89 \\

        Ours w/o II
        & \colorbox{bestcolor}{22.90}
        & \colorbox{secondcolor}{0.9178}
        & \colorbox{secondcolor}{0.0534}
        & \colorbox{secondcolor}{78.30}
        & \colorbox{secondcolor}{8.176}
        & \colorbox{secondcolor}{36.94}
        \\

        Ours
        & \colorbox{bestcolor}{22.90}
        & \colorbox{bestcolor}{0.9179}
        & \colorbox{bestcolor}{0.0532}
        & \colorbox{bestcolor}{78.38}
        & \colorbox{bestcolor}{7.992}
        & \colorbox{bestcolor}{36.47}
        \\

        \bottomrule
    \end{tabular}%
}

\vspace{2pt}
\parbox{\columnwidth}{\scriptsize \textit{Note:}
        Ours w/o II denotes only vehicle-domain adaptation without using driving-log-robust adaptation.
        $\mathrm{CLIP}_{\mathrm{img}}$ is 
reported $\times100$. AH denotes Asset Harvester~\cite{asset_harvester}. 
        We highlight the \colorbox{bestcolor}{best} and \colorbox{secondcolor}{second-best} results.
}
\vspace{-4mm}
\endgroup
\end{table}

\noindent\textbf{Adaptation analysis. }
Table~\ref{tab:vehicle_generation} also compares unadapted TRELLIS,
Stage~I alone (Ours w/o II), and the full model.
Stage~I reduces $\mathrm{FD}_{\mathrm{incep}}$ from $9.788$
to $8.176$; Stage~II further reduces it to $7.992$, with
additional gains in SSIM, LPIPS, $\mathrm{CLIP}_{\mathrm{img}}$,
and $\mathrm{FD}_{\mathrm{dinov2}}$ at unchanged PSNR.
Stage~I provides most of the gains on held-out RECAR, consistent with
adapting the general 3D prior to vehicle geometry and appearance.
Stage~II keeps the pseudo-targets fixed but perturbs the conditioning
image, training the model to match the same pseudo-targets from
less reliable observations. Its smaller gains on clean RECAR views
and fewer visible appearance errors in the real driving-log examples
in Fig.~\ref{fig:actor_qualitative} are consistent with reduced sensitivity
to input perturbations. This distinction explains why the clean-image
benchmark alone gives an incomplete picture of Stage~II's role.
Together, these results support RECAST’s ability to generate 
view-complete vehicles from driving-log observations while 
retaining appearance cues from the input.

\section{Conclusion}
We presented RECAST, a framework that integrates view-complete vehicle actors generated 
from single driving-log images into reconstructed scenes for controlled planner-in-the-loop 
simulation. Two-stage adaptation on RECAR improves actor generation quality, while scene-level 
experiments show improved rendering realism over native actors beyond logged trajectories. 
Closed-loop experiments with a fixed GTRS-Dense planner show that outcomes depend on the 
rendered observations as well as the planner, highlighting the importance of actor rendering 
in simulation-based evaluation.

\noindent\textbf{Future work. }We plan to extend RECAST to pedestrians and other traffic actors, 
enabling more complex interactions in simulation. We also aim to  
use controllable traffic interactions 
to  train  driving  planners,  with  an  emphasis  on  rare  and 
safety-critical scenarios. We will assess whether this training 
improves robustness and transfers to real-world driving.

\end{document}